%% file: main.tex
\documentclass[10pt,twocolumn,letterpaper]{article}

\usepackage{cvpr}              
\input{preamble}
\definecolor{cvprblue}{rgb}{0.21,0.49,0.74}
\usepackage[pagebackref,breaklinks,colorlinks,allcolors=cvprblue]{hyperref}

\usepackage{graphicx}
\usepackage{amsmath}
\usepackage{amssymb}
\usepackage{enumitem}
\usepackage{booktabs}
\usepackage{adjustbox}
\usepackage[table]{xcolor}
\usepackage{units}
\usepackage{multirow}
\usepackage{makecell}
\usepackage{mathtools}

\usepackage{xcolor}

\usepackage{utfsym}

\colorlet{alternateRowColor}{teal!5}

\newcommand{\coloredBelowRuleSep}[1]{
    \arrayrulecolor{#1}
    \specialrule{\belowrulesep}{0pt}{0pt}
    \arrayrulecolor{black}
}

\newcommand{\coloredMidrule}[2]{
    \arrayrulecolor{#1}
    \specialrule{\aboverulesep}{0pt}{0pt}
    \arrayrulecolor{black}
    \specialrule{\lightrulewidth}{0pt}{0pt}
    \coloredBelowRuleSep{#2}
}

\newcommand{\coloredBottomrule}[1]{
    \arrayrulecolor{#1}
    \specialrule{\aboverulesep}{0pt}{0pt}
    \arrayrulecolor{black}
    \specialrule{\heavyrulewidth}{0pt}{0pt}
    \coloredBelowRuleSep{white}
}

\def\paperID{} 
\def\confName{CVPR}
\def\confYear{2026}

\title{Sensitivity as a Double-Edged Sword: \\ A Trade-off Between Discriminability and Adversarial Robustness}

\author{Kai Wang\\
{\tt\small kkaiwwana@gmail.com}}

\begin{document}
\maketitle
\input{tex/0_abstract}
\input{tex/1_introduction}
\input{tex/2_related_work}
\input{tex/3_our_method}

\input{tex/4_experiment}
\input{tex/5_conclusion}

{
    \small
    \bibliographystyle{ieeenat_fullname}
    \bibliography{main_bib}
}

\input{supplementary}

\end{document}

%% file: tex/0_abstract.tex
\begin{abstract}
Modern neural networks are highly susceptible to adversarial perturbations. In this work, we identify that part of this vulnerability stems from the sensitivity of the widely used fully connected (FC) classifiers to such perturbations. In contrast, simple $\ell_2$ distance-based classifiers exhibit significantly greater robustness. We provide thorough theoretical and empirical analysis showing that while FC classifiers' high sensitivity makes them discriminative, it also makes them vulnerable. Conversely, $\ell_2$-classifiers' insensitivity grants robustness but limits performance.
Motivated by this trade-off, we propose a novel $\ell_2$-reclassifier based on a Hybrid Prototype Mixing (HPM) framework. This method retains the discriminative power of FC classifiers while leveraging the robustness of $\ell_2$ distance. It yields $\ell_2$-distance-based predictions by fusing two prototype types: (1) stable, dataset-level prototypes updated via EMA, and (2) dynamic, batch-level prototypes generated from the FC classifier's predictions using a Straight-Through Estimator (STE).
However, this dynamic, STE-based architecture introduces significant challenges for evaluation, such as gradient obfuscation and forward discontinuity. To address this, we propose a new, rigorous evaluation protocol, the Mixed Surrogate Attack (MSA), which uses multiple surrogates along with powerful AutoAttack to ensure a fair and robust assessment. Extensive experiments demonstrate that our lightweight, plug-and-play module, with minimal fine-tuning, effectively enhances the adversarial robustness of various existing SOTA adversarially trained models.
\end{abstract}

%% file: tex/1_introduction.tex
\begin{figure}[t]
\centering
\includegraphics[width=\linewidth]{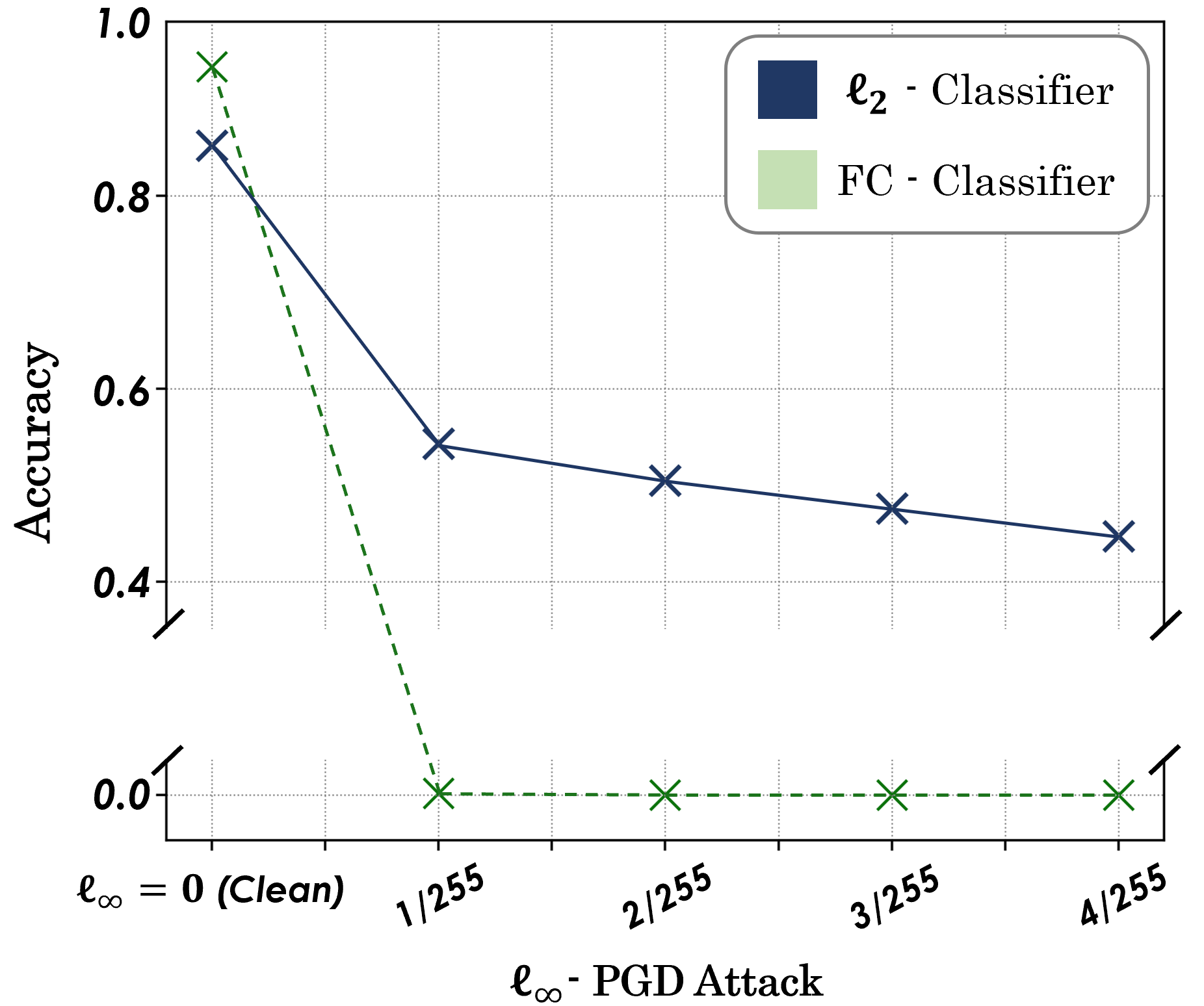}
\caption{Accuracy on CIFAR-10~\cite{krizhevsky2009cifar} validation set under APGD attack from AutoAttack~\cite{autoattack}). We adopt ResNet-34~\cite{he2016deep} to CIFAR-10 and replace its classifier with a simple $\ell_2$-classifier.
}
\label{fig:robust}
\end{figure}

\section{Introduction}
\label{intro}
Modern neural networks are highly vulnerable to adversarial perturbations~\cite{madry2017towards}. Numerous studies have investigated the causes of this phenomenon, such as the local linearity of neural networks~\cite{goodfellow2014explaining}, reliance on non-robust features~\cite{ilyas2019adversarial}, and limited out-of-distribution generalizability~\cite{tsipras2019robustness}. In this paper, we offer a new perspective on this vulnerability, focusing on the classifier component.
We first observe that a standard FC classifier exhibits significantly weaker adversarial robustness compared to a simple $\ell_2$-distance-based classifier (see Fig. \ref{fig:robust}). Under a minimal $\ell_\infty$ perturbation, a model with an FC classifier's accuracy can collapse, while an $\ell_2$-classifier maintains high accuracy, even when both models are trained normally without adversarial training. This highlights the inherent robustness of the $\ell_2$-classifier.

To understand this discrepancy, we analyze both classifiers from a unified perspective, focusing on their sensitivity to feature variations. Our theoretical and empirical analysis ($\S$\ref{sec:sensi_diff}-\ref{sec:dist_analysis}) reveals that FC classifiers are almost always more sensitive to feature variations than $\ell_2$-classifiers, especially in high-dimensional spaces. This sensitivity is a double-edged sword: it allows FC classifiers to capture subtle, discriminative differences in clean data, but it also makes them highly susceptible to adversarial noise. Conversely, the $\ell_2$-classifier's insensitivity reduces the impact of adversarial noise but hinders its ability to capture meaningful differences between classes ($\S$\ref{sec:double_edge_sword}).

Motivated by this, we aim to design a solution that leverages the robustness of $\ell_2$-classifiers while retaining the discriminative power of FC classifiers. To this end, we propose a Hybrid Prototype Mixing (HPM) framework ($\S$\ref{sec:hybrid_model}). This $\ell_2$-reclassifier generates adaptive prototypes by fusing stable, dataset-level centroids (updated via EMA) with dynamic, batch-level centroids. Crucially, these dynamic centroids are generated based on the predictions of the existing FC classifier, using a Straight-Through Estimator (STE) to ensure end-to-end trainability. The final classification is then performed using the robust $\ell_2$-distance to these fused hybrid prototypes.

However, this dynamic, input-dependent architecture, which relies on the discrete argmax operation (via STE), poses a significant challenge to standard adversarial evaluation. It creates issues of gradient obfuscation and forward discontinuity, where small input perturbations can cause abrupt changes in the model's internal state, rendering standard gradient-based attacks ineffective ($\S$\ref{sec:evaluation_protocol}). A naive evaluation would thus report a false sense of security.

To fairly assess our defense, we introduce as a second key contribution: a new, rigorous evaluation protocol called the Mixed Surrogate Attack (MSA) ($\S$\ref{sec:evaluation_protocol}). This protocol is specifically designed to evaluate dynamic, discontinuous defenses by leveraging an ensemble of white-box surrogate models (including the base FC classifier itself) to find worst-case adversarial examples.

We demonstrate the effectiveness of our HPM reclassifier through extensive experiments ($\S$\ref{sec:experiment}) on CIFAR-10, CIFAR-100, and ImageNet. We show that our module, when applied as a lightweight, plug-and-play component to existing SOTA adversarially trained models, can further enhance their robustness with minimal, classifier-only fine-tuning.
In short, our main contributions are:
\vspace{0.3em}
\begin{itemize}[leftmargin=*]
\item We explore the fundamental trade-off between sensitivity, discriminative power, and adversarial robustness by comparing FC and $\ell_2$-classifiers.

\item We propose a novel $\ell_2$-reclassifier, the Hybrid Prototype Mixing (HPM) framework, that leverages FC predictions to create robust, dynamic $\ell_2$-decision boundaries.

\item We identify the evaluation challenges (gradient obfuscation, discontinuity) inherent in such dynamic defenses and propose a rigorous Mixed Surrogate Attack (MSA) protocol to address them.

\item We demonstrate through extensive experiments that our method effectively improves the robustness of existing SOTA models via minimal fine-tuning.
\end{itemize}

%% file: tex/2_related_work.tex
\section{Related Work}
\label{related_work}

\noindent \textbf{Adversarial Attack and Defense.}
Adversarial attack and defense is a critical area, in enhancing the robustness and security of AI systems, which are vulnerable~\cite{szegedy2013intriguing,michel2022survey_vulnerale, bortsova2021adversarial_vulnerable} to adversarial samples which are subtly altered inputs~\cite{szegedy2013intriguing, goodfellow2014explaining} (the strength of perturbations are usually constrained by $\ell_p$ norm, e.g. $\ell_2, \ell_\infty$) to deceive models. Research in this field primarily considers two settings: black-box and white-box. In black-box attacks~\cite{papernot2017practical, chen2017zoo}, attackers have limited access, typically only interacting with the model’s outputs, while white-box attacks allow full access to model internals, enabling more optimized attack strategies. Key algorithms have set milestones in this field. For example, the fast gradient sign method (FGSM)~\cite{goodfellow2014explaining} and projected gradient descent (PGD)~\cite{madry2017towards} are prominent white-box attack methods, while adversarial training~\cite{goodfellow2014explaining, madry2017towards, bai2021recent} remains one of the most effective defense mechanisms, hardening models by exposing them to adversarial examples during training. In this work, we will expose models armored by our $\ell_2$-reclassifier to various kinds of attack algorithms, using standardized benchmark RobustBench~\cite{croce2020robustbench}.

\noindent \textbf{Distance-based Classifiers.}
\label{related_work_cls}
Early distance-based (DB) classifiers, such as K-Nearest Neighbors (KNN)~\cite{cover1967nearest}, Minimum Distance Classifier (MDC)~\cite{hodgson1988min_dis_classifier}, and Nearest Class Mean (NCM)~\cite{webb2003statistical}, date back to the last century. Among them, representing each class by its centroid offers strong interpretability~\cite{wang2022visual} and efficiency. However, traditional DB methods struggle with high-dimensional~\cite{beyer1999nearest, aggarwal2001surprising}, large-scale data. With the advent of deep networks like ResNet~\cite{he2016deep}, which map inputs into lower-dimensional space, DB classifiers have regained attention. This revival has led to methods such as Matching Networks~\cite{vinyals2016matching} and Prototypical Networks~\cite{snell2017prototypical}, particularly in few-shot and zero-shot learning~\cite{wang2020generalizing, romera2015embarrassingly}. Subsequent improvements focus on optimizing prototype representation and incorporating learnable components~\cite{gogoi2022adaptive, ji2020improved, mettes2019hyperspherical, gao2019hybrid}. These techniques have also been extended to domain adaptation~\cite{ben2006analysis, pan2019transferrable} and continual learning~\cite{parisi2019continual, mai2021scr_ncm}. Nevertheless, DB classifiers generally underperform FC classifiers when scaled to larger datasets and class counts. An exception is Deep Nearest Centroids (DNC)~\cite{wang2022visual}, which leverages clustering algorithm for sub-centroids discovery and achieves performance comparable to FC classifiers on ImageNet~\cite{deng2009imagenet}. However, its accuracy only dropped slightly from 77.80\% to 77.31\%, while the FC baseline achieves 77.52\%, when reduced to a single centroid per class (in this case, it has same parameters counts to FC classifier). This suggests that the clustering mechanism is not the key factor behind DNC's success. While the use of a memory bank helps generate more informative centroids, we argue that the critical factor is the use of cosine similarity over normalized features\footnote{Their implementation applies layer normalization~\cite{ba2016layer} and $\ell_2$ normalization to both features and class centroids.}. This design mimics FC classifiers and leads to comparable results. In contrast, DeepNCM~\cite{guerriero2018deepncm}, a similar work that leverages $\ell_2$-classifier, experiences severe performance decay on larger CIFAR-100~\cite{krizhevsky2009cifar} dataset.

\noindent\textbf{Deep Metric Learning.}
Despite what its name might suggest, most work in deep metric learning (DML)~\cite{kaya2019deep} does not aim to explicitly design an interpretable, task-specific distance function~\cite{musgrave2020metric}. Instead, the primary focus has been on learning a better embedding space~\cite{schroff2015facenet}, where intra-class samples are pulled closer and inter-class samples are pushed farther apart. As such, although some recent studies~\cite{mao2019metric} have explored the use of DML for adversarial defense, their emphasis lies in leveraging triplet-based losses~\cite{hoffer2015deep} to obtain more robust feature representations, rather than altering the form of the decision function~\cite{weinberger2009distance} itself, which remains structurally similar to a FC layer operating under cosine distance. Instead, to the best of our knowledge, our work is the first to identify that simply changing the decision rule can lead to substantial differences in adversarial robustness.

%% file: tex/3_our_method.tex
\section{Methodology}
\label{method}
\subsection{Notations}

Let $x \in \mathbb{R}^{h \times w \times c}$ and $X \in \mathbb{R}^{b \times h \times w \times c}$ denote a single image and a batch of images, respectively, where $b$ is the batch size, $c$ the number of channels, and $h, w$ the image height and width. The true and predicted labels are denoted by $y, \hat{y} \in \{0, \dots, k{-}1\}$ (or $\{0, \dots, k{-}1\}^b$ for batches), where $k$ is the number of classes. We denote the backbone (feature extractor) as $\phi$, which maps $x$ or $X$ to a feature vector $z \in \mathbb{R}^d$ or $Z \in \mathbb{R}^{b \times d}$, where $d$ is the feature dimension. All vectors are assumed to be row vectors by default.

\subsection{Unified Perspective of Classifiers}
\label{fc_or_dist}
The fundamental difference between a DB classifier and a FC classifier lies in their respective decision and optimization forms. We first define the DB classifier $F_d$ and the FC classifier $F_c$, both of which map feature representations to output logits. For simplicity and considering the effectiveness of multi-centroids strategy, we only discuss DB classifiers with a single centroid per class. Let $W^d \in \mathbb{R}^{k \times d}$ and $W^c \in \mathbb{R}^{d \times k}$ denote the weight matrices of the DB and FC classifiers (excluding bias), respectively. The FC classifier predicts labels via:
\begin{equation} \label{eq:1}
    \hat{y}_c = \mathrm{argmax}~(z \cdot W^c) = \mathrm{argmax}~[\langle z, W^c_{:,i} \rangle]_{i=0}^{k-1}
\end{equation}
where $W^c_{:,i}$ denotes the $i$-th column of $W^c$. In contrast, the DB classifier is defined as:
\begin{equation} \label{eq:2}
    \hat{y}_d = \mathrm{argmax}~-d(z, W^d) = \mathrm{argmax}~-[d(z, W^d_i)]_{i=0}^{k-1}
\end{equation}
where $d(\cdot,\cdot)$ is a distance metric and $W^d_i$ is the centroid of class $i$. Interestingly, in recent SOTA work~\cite{wang2022visual}, the DB classifier uses cosine similarity between normalized features and centroids. Under this setting, the DB classifier becomes equivalent in form to the FC classifier, and therefore yields comparable performance, whereas the solution with $\ell_2$-classifier~\cite{guerriero2018deepncm} fails. This implies that the two classifiers can be made nearly equivalent through decision form design, yet may differ substantially when their decision mechanisms diverge. Regarding optimization, modern DB classifiers~\cite{wang2022visual, guerriero2018deepncm, mai2021scr_ncm} typically learn the feature extractor jointly with classification, rather than fitting centroids on fixed features as in early works~\cite{mensink2013distance}. In this setting, the centroids are not explicitly updated via gradient descent, yet implicitly shaped by supervised learning through the features. So what truly differentiates the two? We identify a key insight: compared to FC classifiers, those using $\ell_2$-distance tend to be less sensitive to variations in feature representations. As we will show, this insensitivity is a double-edged sword, providing improved robustness but reduced discriminative capability.

\subsection{Sensitivity to Feature Variations}
\label{sec:sensi_diff}
We provide a simple insight: DB classifiers, specifically those using $\ell_2$-distance, tend to be less sensitive to feature perturbations than FC classifiers based on inner products. For clarity, we assume that the variation in feature representation is $\epsilon \in \mathbb{R}^d$, and both the feature vector $z$ and the perturbation $\epsilon$ have the same $\ell_2$ norm under both classifiers, as they share the same feature extractor. Let $w_c$ and $w_d$ denote the reference vectors for the FC and DB classifiers (i.e., $W^c_{:,i}$ and $W^d_i$), respectively. The relative sensitivity of the DB classifier, denoted as $\mathbf{S}_c \in \mathbb{R}_{\geq 0}$, is defined as:
\begin{align}
\label{eq:r_c}
        \mathbf{S}_c = \left|\frac{(z_c + \epsilon_c)\cdot w_c - {z_c\cdot w_c}}{z_c\cdot w_c}\right| = \frac{\|\epsilon_c\|\cdot |\mathrm{cos(\gamma)}|}{\|z_c\|\cdot |\mathrm{cos}(\theta_c)|}
\end{align}
where $\theta_c$ is the angle of $z_c$ and $w_c$, and $\gamma$ is the angle of $\epsilon_c$ and $w_c$. Similarly, the relative sensitivity for $\ell_2$-classifier, denoted as $\mathbf{S}_d \in \mathbb{R}_{\geq 0}$, is:
\begin{align}
\label{eq:r_d}
    \mathbf{S}_d &=\left|\frac{\|(z_d + \epsilon_d) -w_d\|- \|z_d-w_d\|}{\|z_d-w_d\|}\right|
\end{align}
We aim to explore the relationship between $\mathbf{S}_c$ and the supremum of $\mathbf{S}_d$. First, we can easily derive the supremum of $\mathbf{S}_d$ via triangle inequity and cosine formula, which is given as follows:
\begin{align}
\label{eq:sup_r_d}
    \mathrm{sup}~\mathbf{S}_d & = \frac{\|\epsilon_d\|}{\sqrt{\|z_d\|^2+\|w_d\|^2-2\|z_d\|\|w_d\| \mathrm{cos}(\theta_d)}}
\end{align}
where $\theta_d$ is the angle of $z_d$ and $w_d$. Then, we can derive an equivalent form and further give a sufficient condition for $\mathbf{S}_c$ being greater or equal to $\mathrm{sup}~\mathbf{S}_d$ as follows:
\begin{align}
\label{eq:infer}
    \mathbf{S}_c \geq& ~\mathrm{sup}~\mathbf{S}_d \iff\\
    |\mathrm{cos}(\theta_c)|\leq& |\mathrm{cos}(\gamma)|\cdot \sqrt{\left(\frac{\|w_d\|}{\|z_c\|}\right)^2-2\cdot\mathrm{cos}(\theta_d)\frac{\|w_d\|}{\|z_c\|}+1} \nonumber \\
    \Longleftarrow~&|\cos(\theta_c)|\leq|\mathrm{cos}(\gamma)|\cdot |\mathrm{sin}(\theta_d)|
\end{align}
Thus, we can derive the following inference, which indicates under what sufficient conditions the decision results based on inner products are more sensitive:
\begin{equation}
\label{eq:conclu}
    \frac{|\mathrm{cos}(\theta_c)|}{|\mathrm{sin}(\theta_d)|}\leq |\mathrm{cos}(\gamma)| \Longrightarrow \mathbf{S}_c \geq ~\mathrm{sup}~\mathbf{S}_d
\end{equation}

\subsection{Distributional Analysis}
\label{sec:dist_analysis}
Next, we are interested in the distribution of random variable \(\nicefrac{|\mathrm{cos}(\theta_c)|}{|\mathrm{sin}(\theta_d)|}\). From this, we can infer the probability that Eq.\eqref{eq:conclu}'s condition holds. To proceed, we denote \(\mathcal{H} = \cos(\theta_c)\) and \(\mathcal{T} = \sin(\theta_d)\), and their respective probability density functions (PDFs) are denoted as \(p_\mathcal{H}\) and \(p_\mathcal{T}\). Next, we aim to derive the cumulative distribution function (CDF) $F_\mathcal{X}(x)$ of \(\mathcal{X} = \left| \nicefrac{\mathcal{H}}{\mathcal{T}} \right|\), with the process as follows:
\begin{align}
    F_\mathcal{X}(x)&=\mathrm{P}(\mathcal{X}\leq x)=\mathrm{P}(|\mathcal{\frac{H}{T}}|\leq x)=\mathrm{P}(|\mathcal{H}|\leq \mathcal{T}\cdot x) \nonumber\\
    &=\int_{0}^1 p_\mathcal{T}(\tau, d)\cdot\mathrm{P}(|\mathcal{H}|\leq \tau\cdot x)~\mathrm{d}\tau \nonumber \\
    &=\int_{0}^{1} p_\mathcal{T}(\tau,d)
    \int_{-\tau\cdot x}^{\tau\cdot x}p_\mathcal{H}(\eta,d)\mathrm{d}\eta\mathrm{d}\tau
    \label{eq:abs_cdf}
\end{align}

\begin{figure}[t]
    \centering
    \includegraphics[width=0.48\textwidth]{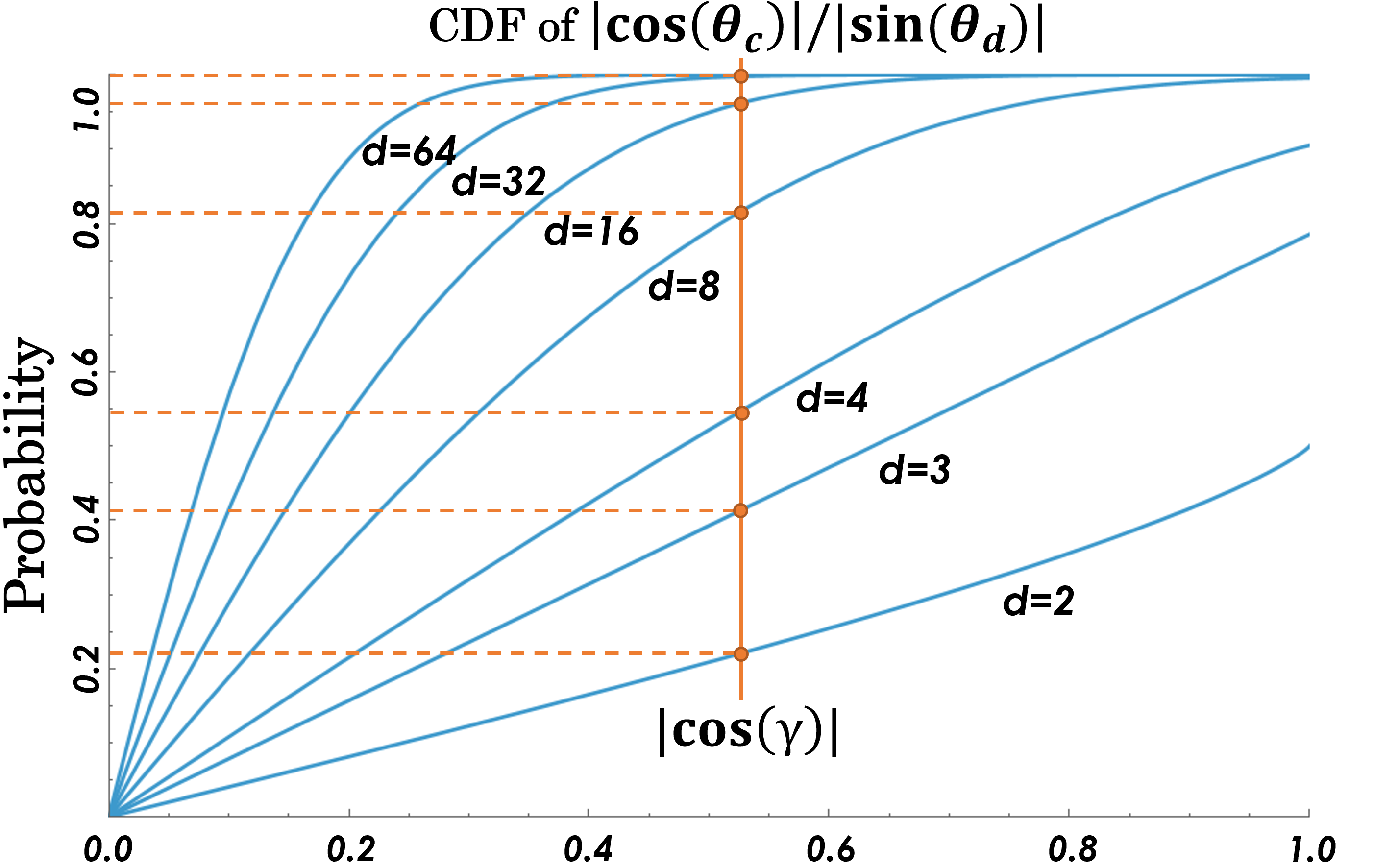}
    \caption{The CDFs of $\mathcal{X}$ with given dimensions under isotropic assumption. Notably, $\gamma$ is the angle between feature variation and reference vector in FC classifier, which tends to be near zero ($|\cos(\gamma)|$ would be close to 1.0) to results in greater perturbations under adversarial attack.}
    \label{fig:cdf}
\end{figure}

\noindent \textbf{Isotropic Distribution.} If we assume that $z$ is uniformly distributed in direction over the space, or alternatively adopt a more conveniently defined concept, we assume that $z$ follows an isotropic distribution, e.g., multivariate Gaussian distribution with a shared variance in each component. These two assumptions are equivalent when the norm of $z$ is fixed~\cite{papaspiliopoulos2020high}; otherwise, isotropy is a sufficient condition for directional uniformity. Then, basing on angle distribution in random packing on sphere~\cite{cai2013distributions}, $p_\mathcal{H}$ and $p_\mathcal{T}$ can be easily derived as follows (the proof can be found in the supplementary material):
\begin{align}
\begin{cases}
    p_\mathcal{H}(\eta, d)= \frac{\Gamma(\nicefrac{d}{2})}{\Gamma(\frac{d-1}{2})\sqrt{\pi}}\cdot(1-\eta^2)^{\frac{d-1}{2}-1}     \\
    p_\mathcal{T}(\tau, d)= \frac{2\Gamma(\nicefrac{d}{2})}{\Gamma(\frac{d-1}{2})\sqrt{\pi}}\cdot \frac{\tau^{d-2}}{({1-\tau^2})^{\nicefrac{1}{2}}}
    \label{eq:istropic_pdfs}
\end{cases}
\end{align}
Then, the probability that Eq.\eqref{eq:conclu} holds can be derived by combining Eq.\eqref{eq:abs_cdf} and Eq.\eqref{eq:istropic_pdfs}, as follows (also see Fig.~\ref{fig:cdf}):
\begin{align}
\label{eq:conclu_prob}
    &\mathrm{P}(\frac{|\mathrm{cos}(\theta_c)|}{|\mathrm{sin}(\theta_d)|}\leq |\mathrm{cos}(\gamma)| ) = F_{\mathcal{X}}(|\mathrm{cos}(\gamma)|)=\\
    &\frac{2\cdot\Gamma^2(\frac{n}{2})}{\pi\cdot\Gamma^2(\frac{n-1}{2})}\int_0^{1}\frac{x^{d-2}}{\sqrt{1-x^2}}\int_0^{\substack{x\cdot\mathrm{cos}(\gamma)}}(1-y^2)^{\frac{n-3}{2}}\mathrm{d}x\mathrm{d}y \nonumber
\end{align}

\noindent \textbf{Empirical Distribution.} 
Although the above results indeed characterize that, as the feature dimensionality increases, the FC classifier is almost always more sensitive than the $\ell_2$-classifier, in practical scenarios, the input feature vectors are rarely randomly distributed over the entire space. Instead, they are typically correlated with the reference vectors of each class in the classifier. However, we cannot analytically derive the distribution of the angles between them. Therefore, we aim to model the real features and classifier parameters from large-scale datasets and, based on this modeling, obtain an analytically tractable empirical probability distribution. Specifically, we extract features from the dataset using a trained network and statistically analyze the angular distributions between these features and the reference vectors in the classification head. In particular, considering the distinct relationships between a feature vector and its corresponding class versus non-corresponding class reference vectors, we model these angles separately as positive and negative samples. Formally, for a feature vector $z$ of class $K$ and reference vectors ${w_d^{i=K}}$ and ${w_d^{i \neq K}}$, we respectively compute the distributions of their angles and fit them using truncated Gaussian distributions. The resulting PDFs are then combined according to the sample sizes to obtain the final empirical distribution. Since we are dealing with angular distributions, we truncate the Gaussian range to $[0, \pi]$ and estimate the empirical parameters $\mu_c^+$, $\sigma_c^+$ and $\mu_c^-$, $\sigma_c^-$ accordingly. Similarly, we obtain the empirical parameters $\mu_d^+$, $\sigma_d^+$ and $\mu_d^-$, $\sigma_d^-$ for the trained distance-based classifier. Then the PDF of $\theta_d$ and $\theta_c$ are given as follows:
\begin{align}
    \begin{cases}
            p_{\theta_c}(x)= \alpha_c \cdot t(x,\mu_c^+,\sigma_c^+) + (1-\alpha_c)\cdot t(x,\mu_c^-,\sigma_c^-)\\
            p_{\theta_d}(x)= \alpha_d \cdot t(x,\mu_d^+,\sigma_d^+) + (1-\alpha_d)\cdot t(x,\mu_d^-,\sigma_d^-) \\
            t(x,\mu,\sigma)=\frac{1}{\sigma}\cdot\frac{\varphi(\nicefrac{(x-\mu)}{\sigma})}{\Phi(\nicefrac{(\pi-\mu)}{\sigma})+\Phi(\nicefrac{\mu}{\sigma})-1}
    \end{cases}
    \label{eq:emp_dist}
\end{align}
where $\alpha$ is the ratio of number of samples with corresponding category, $\varphi(\cdot)$ is the PDF of standard normal distribution and $\Phi(\cdot)$ is its CDF. By applying variable substitution, we can easily derive $p_\mathcal{H}$ and $p_\mathcal{T}$, and then empirical probability can be derived as follows:
\begin{align}
    \label{eq:modeled_cdf}
    &\mathrm{P}(\frac{|\mathrm{cos}(\theta_c)|}{|\mathrm{sin}(\theta_d)|}\leq |\mathrm{cos}(\gamma)| ) = F_{\mathcal{X}}(|\mathrm{cos}(\gamma)|)=\\
    &\int_0^{\frac{\pi}{2}}{\left(p_{\theta_d}(x)+p_{\theta_d}(\pi-x)\right)}\int_{\substack{\mathrm{sin}(x)\cdot\\-\mathrm{cos}(\gamma)}}^{\substack{\mathrm{sin}(x)\cdot\\\mathrm{cos}(\gamma)}}\frac{p_{\theta_c}(\mathrm{cos^{-1}}(y))}{\sqrt{1-y^2}}\mathrm{d}x\mathrm{d}y \nonumber        
\end{align}


\begin{figure}[b]
    \centering
    \includegraphics[width=\linewidth]{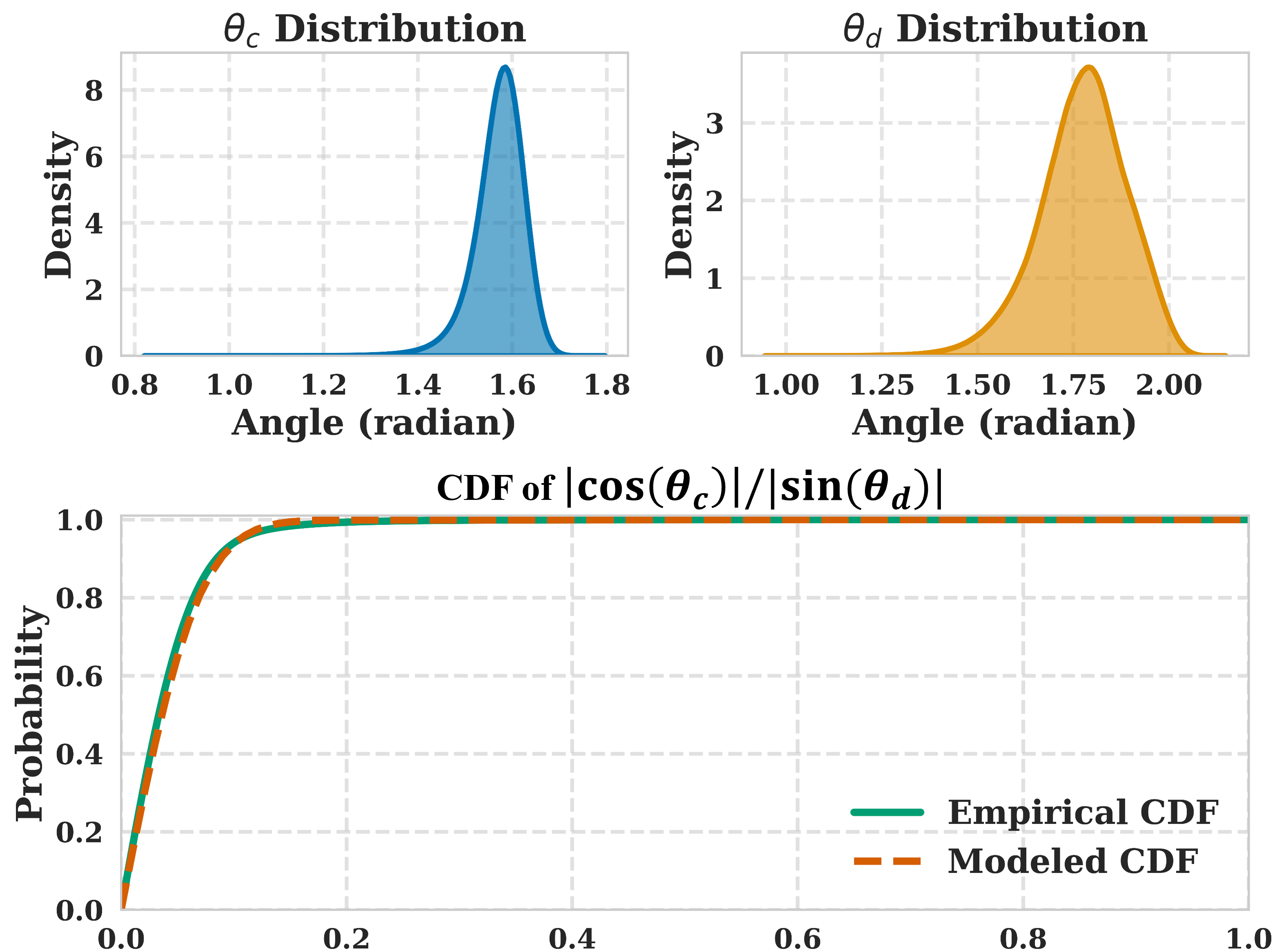}
    \caption{The empirical distribution of $\theta_c$, $\theta_d$ and the empirical and modeled CDF (from Eq.~\eqref{eq:modeled_cdf}) of random variable $\nicefrac{\cos(\theta_c)}{\sin(\theta_d)}$.}
    \label{fig:emp_cdf}
\end{figure}
 
We begin by validating the rationality of our assumption regarding the angular distribution of positive and negative samples on the ImageNet validation set. Specifically, we observe that the angles between feature vectors and their corresponding class prototypes (positive samples), and those with non-corresponding class prototypes (negative samples), form two distinct modes. Using ResNet-50 as an example, the positive sample angles exhibit a mean of $\mu_c^+ = 66.81^\circ$ and a standard deviation of $\sigma_c^+= 5.56^\circ$, while the negative sample angles are highly concentrated around $90^\circ$ (with $\mu_c^- = 90.01^\circ$ and $\sigma_c^- = 3.02^\circ$). Similarly, we train a ResNet-50 equipped with a $\ell_2$-classifier on ImageNet. The resulting angular statistics, denoted as $\mu_d^+$ and $\sigma_d^+$ for positives and $\mu_d^-$ and $\sigma_d^-$ for negatives, are $74.3^\circ$ and $7.9^\circ$, and $101.3^\circ$ and $4.3^\circ$, respectively. By modeling the empirical distributions as in Eq.\eqref{eq:emp_dist}, and substituting the observed statistics $\mu_c^+, \sigma_c^+$ and $\mu_c^-, \sigma_c^-$, we can derive an approximated CDF for the probability in Eq.\eqref{eq:conclu}. The corresponding results are illustrated in Fig.\ref{fig:emp_cdf}, showcasing our method yields a plausible analytical model for such distribution compared with observed empirical distribution. These findings collectively suggest that, under both idealized and empirical angular distributions, $\ell_2$-classifiers exhibits lower sensitivity to feature variations. This reduced sensitivity contributes to improved robustness; however, it also comes at a cost, as insensitivity to feature variations may hinder discriminability.

\subsection{Sensitivity: A Double-edge Sword}
\label{sec:double_edge_sword}
Through analysis and numerical computation, we can observe that, especially in high-dimensional spaces, decision-making via FC methods is very likely to be more sensitive than using $\mathit{\ell_{2}}$-distance. This trait is essentially a double-edged sword: in the case of adversarial attacks, $\epsilon$ in the above derivation can represent the maximum perturbation that an attacker can apply on the feature space (which can be bounded by the $\mathit{\ell_{2}}$-norm by assumption), and the above conclusion suggests that $\ell_2$-classification strategies are less susceptible to perturbations, meaning such models will have stronger adversarial robustness. However, if $\epsilon$ is viewed as the natural feature differences between samples in the dataset, it implies that $\ell_2$-classifier is less able to capture these differences, leading to weaker classification capabilities. This is consistent with the observations people have made over time, e.g., the performance of DeepNCM that uses a regular $\ell_2$-classifier drops in more challenging datasets, while its close counterpart DNC that leverages inner-product similarity achieves comparable results to FC classifier. In this case, we aim to acknowledge the weaknesses of vanilla $\ell_2$-classifier while leveraging their valuable aspects. Rather than leaving them to operate in isolation, we explicitly take advantage of the merits of FC classifiers. By combining estimated sample class centroids with dataset-level class centroids, we use fused prototypes to achieve refined classification outcomes.

\subsection{Hybrid Prototype Re-Classification Strategy}
\label{sec:hybrid_model}

As discussed in $\S$\ref{sec:double_edge_sword}, the $\ell_2$-classifier’s insensitivity provides robustness at the cost of discriminative precision, whereas the FC classifier exhibits the opposite trade-off. Our primary objective is to devise a learning framework that effectively integrates the strengths of both. Specifically, we aim to exploit the FC classifier’s fine-grained predictive capability to generate dynamic prototypes, and then employ the $\ell_2$-classifier’s robust distance metric to reconstruct a more stable decision boundary.

However, our preliminary experiments revealed a critical limitation in this idea: a model relying solely on dynamic batch prototypes (derived from FC logits) fails to achieve genuine robustness. The apparent defense effect is merely an artifact of the FC classifier’s $\mathrm{argmax}$ dependency. This dependence introduces unavoidable~\footnote{In fact, although we can theoretically replace the $\mathrm{argmax}$ with a fully differentiable low-temperature softmax, its impact on the gradient exploitability for adversarial attacks is equally fatal.} gradient masking issues (will be elaborated in $\S \ref{sec:evaluation_protocol}$), rendering the FC model itself a perfectly exploitable surrogate for adversarial attacks.

To establish a defense with genuine robustness, we propose a \text{Hybrid Prototype Mixing (HPM)} framework. This strategy eliminates exclusive reliance on the vulnerable FC predictions by introducing stable, dataset-level prototypes as anchors. The HPM framework jointly optimizes the backbone $\phi$, the FC classifier $W^c$, and a prototype mixer $m(\cdot, \cdot)$ parameterized by $\phi_m$. It is constructed upon two distinct types of prototypes:
\vspace{0.3em}

\textbf{Dataset-Level Prototypes ($\mathcal{C}_{\text{global}}$).}
We maintain a set of stable dataset-level prototypes $\mathcal{C}_{\text{global}} \in \mathbb{R}^{k \times d}$ as a persistent buffer. Since the feature distribution $\phi(X)$ shifts during training, computing dataset-wide centroids at each step is infeasible. Instead, we update $\mathcal{C}_{\text{global}}$ using an Exponential Moving Average (EMA), aligning with the fact that newer features better reflect the current model state. The update rule, applied during training, is:
\begin{equation}
    \mathcal{C}_{\text{global}} \leftarrow \lambda \cdot \mathcal{C}_{\text{global}} + (1-\lambda) \cdot \mathcal{C}_{\text{gt\_batch}}
    \label{eq:ema_update}
\end{equation}
where $\lambda$ is the momentum coefficient, and $\mathcal{C}_{\text{gt\_batch}}$ is the batch-level prototype computed using ground-truth labels $y$, ensuring the stability and correctness of this anchor.

\vspace{0.3em}
\textbf{Dynamic Batch-Level Prototypes ($\mathcal{C}_{\text{batch}}$).}
We compute dynamic batch-level prototypes $\mathcal{C}_{\text{batch}} \in \mathbb{R}^{k \times d}$ based on the predicted labels from the FC classifier. This provides a fine-grained, adaptive signal. To ensure end-to-end trainability while using discrete $\mathrm{argmax}$ assignments, we use the Straight-Through Estimator (STE) based on the FC logits $L = Z W^c$:
\begin{align} 
\begin{cases}
    \mathcal{P}_{\text{soft}} = \mathrm{softmax}(L)\\ 
    \mathcal{P}_{\text{hard}} = \mathrm{\text{one-hot}}(\mathrm{argmax}(L)) \\
    \mathcal{P}_{\text{ste}}= \mathcal{P}_{\text{soft}} + \mathrm{\text{stop-grad}}(\mathcal{P}_{\text{hard}} - \mathcal{P}_{\text{soft}})
     \label{eq:ste_batch_center}
\end{cases}
\end{align}
From this, $\mathcal{C}_{\text{batch}}$ can be simply obtained as follows:
\begin{equation}
    \mathcal{C}_{\text{batch}} = \frac{(\mathcal{P}_{\text{ste}})^\top Z}{\left[\sum_{i=0}^{b-1}(\mathcal{P_\text{ste}})_{i,j}\right]_{j=0}^{k-1} + \varepsilon}
\end{equation}
where $Z = \phi(X)$, and $\varepsilon$ is a small constant for numerical stability, and the division is broadcasted.

\noindent\textbf{\\Prototype Fusion and Final Prediction.}
Given the reliable but coarse $C_{\text{global}}$ and the fine-grained but noisy $\mathcal{C}_{\text{batch}}$, we introduce a learnable prototype mixer $m(\cdot, \cdot)$ parameterized by $\phi_m = \{W_m, b_m\}$. We adopt a simple linear transformation to fuse them into the final hybrid prototype $C_{\text{hybrid}}$:
\begin{equation}
    \mathcal{C}_{\text{hybrid}} = m(\mathcal{C}_{\text{global}}, \mathcal{C}_{\text{batch}}) = [\mathcal{C}_{\text{global}}, \mathcal{C}_{\text{batch}}] \cdot W_m + b_m
    \label{eq:hybrid_mixer}
\end{equation}
where $W_m \in \mathbb{R}^{2d \times d}$ and $b_m \in \mathbb{R}^d$ are the learnable parameters of the mixer. The model's final output logits, $L_{\text{hybrid}}$, are the negative $\ell_2$-distances from each feature $Z_i$ to this hybrid prototype:
\begin{equation}
    (L_{\text{hybrid}})_{ij} = -\| Z_i - (\mathcal{C}_{\text{hybrid}})_j \|_2
    \label{eq:hybrid_logits}
\end{equation}
The entire architecture, comprising the backbone $\phi$, the FC classifier $W^c$, and the prototype mixer $\phi_m$, can be trained in an end-to-end manner. The HPM framework is hypothesized to enhance robustness because the final decision (Eq.\ref{eq:hybrid_logits}) no longer depends exclusively on the vulnerable FC logit assignments but is instead stabilized through the EMA-updated $C{\text{global}}$, providing a more reliable and resilient decision boundary.

\begin{figure}[h]
    \centering
    \includegraphics[width=\linewidth]{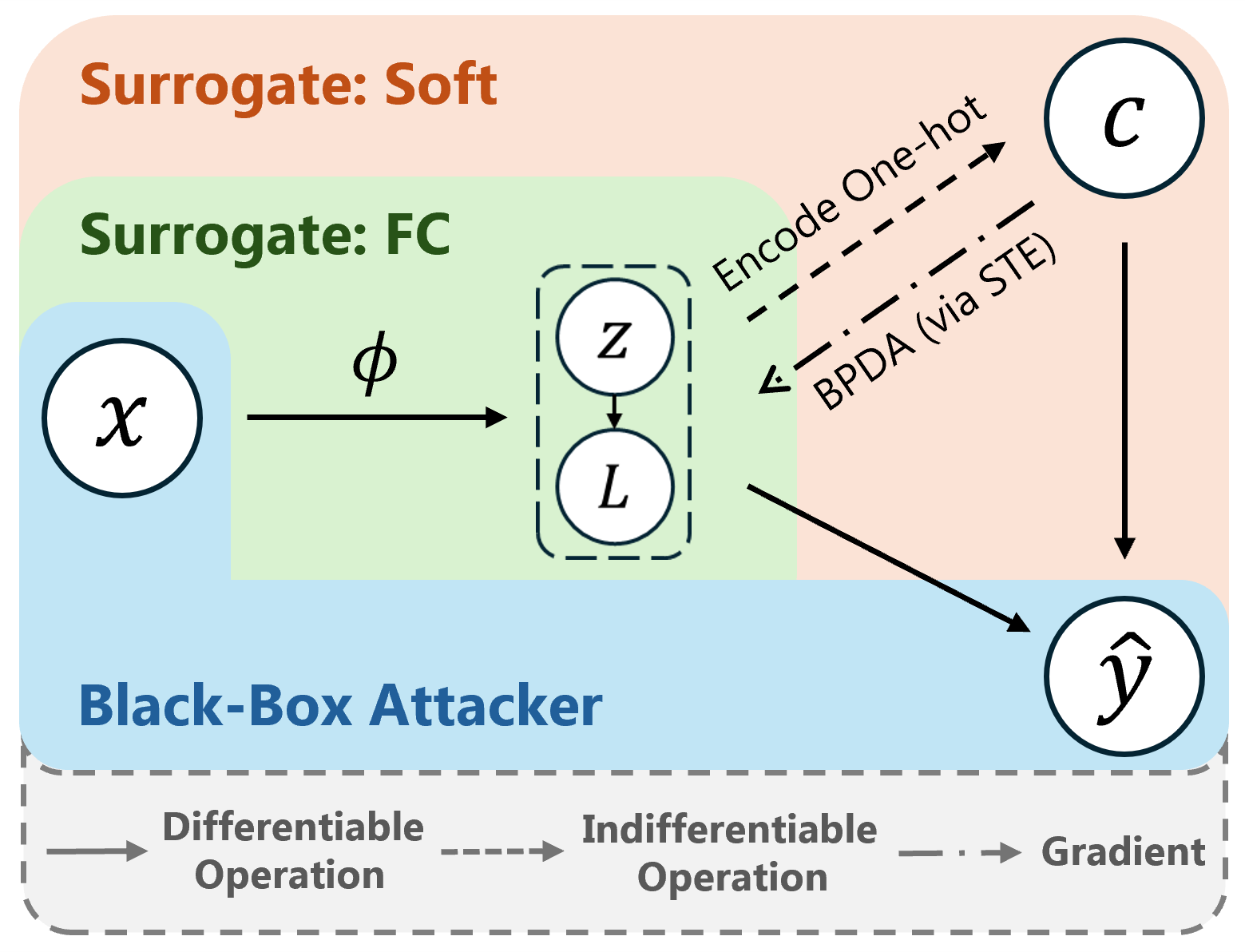}
    \caption{An overview to the proposed evaluation protocol for genuine robustness of dynamic models like our proposed.}
    \label{fig:eval_protocal}
\end{figure}

\subsection{Evaluation Protocol}
\label{sec:evaluation_protocol}
The HPM architecture ($\S \ref{sec:hybrid_model}$), although more robust, still employs a semi-discrete STE mechanism (Eq.\eqref{eq:ste_batch_center}) and therefore poses substantial challenges to standard gradient-based adversarial attacks, resulting in obfuscated gradients~\cite{athalye2018obfuscated}. While such obfuscation may increase the practical difficulty of an attack in real-world, a rigorous academic evaluation requires a worst-case analysis under the assumption of a white-box adversary capable of bypassing these artifacts. The gradient obfuscation observed in our model arises from two principal sources:
\textbf{(1) Gradient Signal Problem:} The STE mechanism supplies a backward gradient (through $\mathcal{P}_{\text{soft}}$) that poorly approximates the true forward mapping, which depends on $\mathcal{P}_{\text{hard}}$.
\textbf{(2) Forward Discontinuity Problem:} Even when a backward gradient is available (e.g., through the built-in STE), the forward pass itself remains discontinuous. A small perturbation $\epsilon$ added to $Z_t$ during the $t$-th step of an attack may cause $\mathrm{argmax}(L_t)$ to switch labels, leading to an abrupt change in the computed $\mathcal{C}_{\text{batch}}$ (Eq.\eqref{eq:ste_batch_center}). Such discontinuities violate the static model assumption of iterative attacks, as the gradient computed at $X_t$ no longer provides a valid direction for minimizing the loss at $X_{t+1}$.

To address these challenges, we first adopt the Backward Pass Differentiable Approximation (BPDA) strategy~\cite{athalye2018obfuscated}. We assume the attacker possesses full knowledge of our architecture, including the STE mechanism. The attacker may thus utilize the smooth gradients derived from $\mathcal{P}_{\text{soft}}$ to perform gradient-based attacks. However, BPDA alone cannot fully resolve the forward discontinuity issue. To establish a stable optimization landscape for the attacker, we introduce a fully soft model that uses $\mathcal{P}_\textbf{soft}$ in forward-pass as our surrogate model. Moreover, considering the fact that such approach base its prediction on FC's, it indicates the vanilla  FC model would be another powerful surrogate model.

Therefore, we propose a \text{Mixed Surrogate Attack (MSA)} protocol that integrates two complementary white-box surrogate models to generate maximally effective adversarial examples:

 \vspace{0.3em}
\textbf{Surrogate 1: The FC Oracle.} This surrogate targets the discontinuity problem directly. The attack is optimized solely against the FC logits, which is the most effective means of inducing abrupt changes in $\mathcal{C}_{\text{batch}}$. This approach poses a genuine threat to our model, as the re-classification process fundamentally relies on FC-based predictions.

 \vspace{0.3em}
\textbf{Surrogate 2: The Soft-Hybrid Model.} This surrogate targets the final $\ell_2$ decision boundary under the assumption of a perfectly smooth STE. It substitutes $\mathcal{P}_{\text{ste}}$ with $\mathcal{P}_{\text{soft}}$ to compute a soft batch centroid $\mathcal{C}_{\text{batch\_soft}}$, which is subsequently fused with the fixed $\mathcal{C}_{\text{global}}$ through the same learned mixer $m(\cdot, \cdot)$.

\noindent \textbf{\\Final Evaluation Protocol.} Our worst-case evaluation is defined as the minimum accuracy observed across a comprehensive suite of attack methods. This protocol is particularly crucial for multi-class scenarios, where both gradient mismatch and decision discontinuity are more pronounced. It is worth noting that, although black-box attacks typically exhibit limited success against conventional models compared to gradient-based attacks, they are of special importance here because they can expose pseudo-robust samples when gradient information is unreliable. Accordingly, our complete evaluation protocol is as follows:

 \vspace{0.3em}
 \textbf{Step 1: Mixed Surrogate Attack (MSA):} An adaptive PGD-based attack (e.g., APGD from AutoAttack) employing an ensemble of surrogates. The generated adversarial examples $X_{\text{adv}}$ are then evaluated on the test model.

 \vspace{0.3em}
 \textbf{Step 2: Gradient-Free Black-Box Attack:} A high-strength black-box attack (Square Attack~\cite{andriushchenko2020square}) on test model to estimate the worst-case robustness independently of any gradient-based assumptions.

\noindent \textbf{\\Rationale and Contributions.}
This evaluation framework is motivated by both practical and academic considerations. From a practical standpoint, dynamic, input-dependent architectures like ours represent a promising direction for developing defenses that inherently challenge standard adversarial methodologies. Our work provides a concrete realization of such a defense. From an academic perspective, this study exposes a critical gap between conventional evaluation protocols and emerging defensive paradigms. It offers a detailed case study of two distinct yet intertwined challenges: gradient mismatch (addressed by BPDA) and forward discontinuity (addressed by surrogate modeling). We hope this work encourages the development of next-generation adaptive attacks designed to rigorously evaluate dynamic, discontinuous, and hybrid defense architectures.

%% file: tex/4_experiment.tex
\begin{table*}[ht!]
    \centering
	\begin{minipage}[t]{0.59\textwidth}
		\centering
            \resizebox{\textwidth}{!}{\input{tabs/adv_robust}}
         \caption{Quantitative results of adversarial robustness experiment on CIFAR-10 and CIFAR-100.}
\label{table:adv_cifar}
	\end{minipage}
	\hfill
	\begin{minipage}[t]{0.37\textwidth}
		\centering
        \resizebox{\textwidth}{!}{\input{tabs/adv_robust_imgnet}}
            \caption{Quantitative results of adversarial robustness experiment on ImageNet.}
    \label{table:adv_imgnet}
	    \end{minipage}
\end{table*}

\begin{table}[ht!]
\centering
\resizebox{0.49\textwidth}{!}{
    \input{tabs/ablation_adv}
}

\caption{Results of ablation study. All setup variants are based on the SOTA model from \cite{xu2023mimir} on ImageNet. Notably, we provide a special implementation and denoted as \textit{FC-Head} in the {DistDeci} column. It means we use FC head again as similarity measure in our reclassification head.} 

\label{table:ablation}
\end{table}

\section{Experiment}
\label{sec:experiment}
\subsection{Datasets} 
\noindent \textbf{CIFAR-10 \& CIFAR-100}
We evaluate the adversarial robustness of models on CIFAR-10 and CIFAR-100 datasets~\cite{krizhevsky2009cifar}. CIFAR-10 has 10 categories of common objects, with 60000 32$\times$32 color images split into 50000 for training and 10000 for testing. While CIFAR-100, with 100 more fine-grained categories, has the same image count but only 600 images per category, making it sparser and more challenging than CIFAR-10.\\

\noindent \textbf{ImageNet} To provide a more credible analysis, we validated our method on a larger-scale dataset, ImageNet~\cite{deng2009imagenet}. However, considering the sheer volume of ImageNet and the high computational cost of generating adversarial examples, all experiments were conducted on a subset of ImageNet. Specifically, we sampled 10 images per class from the training set, resulting in a total of 10,000 images for fine-tuning. For evaluation, we sampled 5 images per class from the validation set, yielding 5,000 images in total.

\subsection{Evaluation Setup} 
We evaluate all models on the integrated standard adversarial robustness benchmark platform, RobustBench~\cite{croce2020robustbench}. To prove our method can further enhance various existing adversarially robust models, including both top-ranked SOTA methods and relatively weaker model. Moreover, since these works typically provide models based on different architectures, we also deliberately choose networks with diverse architectures to validate the applicability of our method. (The specific model architectures can be found in the supplementary materials.) Then, we applied our re-classifier using their pretrained parameters (frozen) and adversarially fine-tune them on the corresponded dataset. This is highly efficient since most of these models were massive, whereas training classifier only is far less demanding. Finally, we report the robust accuracy of the enhanced models under our evaluation protocol. We compare it with the original models’ robust accuracy that we re-evaluated under the same attack settings.  Our implementation details can be found in the supplementary material.


\subsection{Quantitative Results} \label{analysis}
Our main results on CIFAR-10/100 and ImagNet can be found in Tab.~\ref{table:adv_cifar} and Tab.~\ref{table:adv_imgnet}, respectively. It can be observed that our method enhances the robustness of all models we randomly selected. This aligns well with our theoretical analysis: as the number of classes and feature dimensionality increases, the input feature vectors tend to be nearly orthogonal to the reference vectors of almost all categories for FC classifier. This makes them highly susceptible to adversarial perturbations, resulting in large deviations from the original predictions. Moreover, correct predictions, while themselves vulnerable, are further affected by perturbations collaboratively induced by these unrelated classes. In contrast, the re-classifier, which makes decisions based on the $\ell_2$ distance, exhibits a significant advantage. The re-classification strategy also ensures that the distance-based classifier retains strong discriminative power even when dealing with large-scale datasets. We will deeply analyze the effectiveness of each competent in next section.

\subsection{Ablation Study and Analysis} \label{ablation_main_text}
To assess the effectiveness of the proposed method, we decompose it into three components: dataset-level centroid estimation (CEst), the parameterized mixer (PMixer), and the DB decision module (DistDeci). We then remove these components, or when their functionality is indispensable, replace them with the simplest possible alternatives (details can be found in the supplementary material). The results are reported in Tab.~\ref{table:ablation}. All components provide clear and positive contributions to robustness.

We further compare different distance measures to validate the advantage of the $\ell_2$-distance–based classifier. Replacing the negative $\ell_2$ distance with inner-product similarity leads to additional performance degradation (see the last row of Tab.~\ref{table:ablation}). We also examine the vanilla $\ell_2$ classifier. For example, replacing the FC layer of Cui’s model~\cite{cui2305decoupled} with an $\ell_2$ head yields a 4.6\% improvement under the APGD-CE attack on CIFAR-100. However, the vanilla $\ell_2$ classifier fails on ImageNet when applied to Xu’s model~\cite{xu2023mimir}, where the large number of classes amplifies its limited discriminative capacity. This observation is consistent with our analysis.

%% file: tabs/adv_robust.tex
\rowcolors{4}{alternateRowColor}{white}
\begin{tabular}{l|cc|cc}
\rowcolor{white}
\midrule

\multirow{2}{*}{\raisebox{-0.3ex}{\textbf{Works~/~Methods}}}  & \multicolumn{2}{c}{CIFAR-10, $\ell_\infty=\nicefrac{8}{255}$}  & \multicolumn{2}{c}{CIFAR-100, $\ell_\infty=\nicefrac{8}{255}$} \\
\cline{2-5}

 & \raisebox{-0.3ex}{Clean Acc.}  & \raisebox{-0.3ex}{Robust Acc.} & \raisebox{-0.3ex}{Clean Acc.} & \raisebox{-0.3ex}{Robust Acc.} \\

\coloredMidrule{white}{alternateRowColor}
\toprule
\multicolumn{5}{l}{{$\rightarrow \textit{Build on fully-connected classifier}$}} \\
\midrule
Cui~\cite{cui2305decoupled} &    92.9     &    67.7    & 73.1  &   38.6   \\
Wang~\cite{wang2023better} &    93.4     &    71.3      & 73.6 &  42.7   \\

Jia~\cite{jia2022adversarial} &   85.3     & 58.0             & 65.9 &   31.5   \\
Bartoldson~\cite{bartoldson2024adversarial} &    93.7     &   74.1            & -  &   -\\

\coloredMidrule{white}{white}
\toprule
\multicolumn{5}{l}{{$\rightarrow \textit{Build on distance-based classifier} $}} \\
\midrule
Cui~\cite{cui2305decoupled} &    91.8     &   69.8(\textbf{+2.2}) & 72.9 &  39.4(\textbf{+0.8})\\
Wang~\cite{wang2023better} & 92.0 & 72.9(\textbf{+1.6}) & 73.6 & 43.1(\textbf{+0.4})\\

Jia~\cite{jia2022adversarial} &    82.8   & 63.6(\textbf{+5.6})  & 65.9 &  33.3(\textbf{+1.8}) \\
Bartoldson~\cite{bartoldson2024adversarial} &   91.9   &   76.4(\textbf{+2.3}) & -  &   -   \\

\coloredBottomrule{alternateRowColor}
\end{tabular}

%% file: tabs/adv_robust_imgnet.tex
\rowcolors{4}{alternateRowColor}{white}
\begin{tabular}{l|cc}
\rowcolor{white}
\midrule

\multirow{2}{*}{\raisebox{-0.3ex}{\textbf{Works~/~Methods}}}  & \multicolumn{2}{c}{ImageNet, $\ell_\infty=\nicefrac{4}{255}$} \\
\cline{2-3}

 & \raisebox{-0.3ex}{Clean Acc.}  & \raisebox{-0.3ex}{Robust Acc.}\\

\coloredMidrule{white}{alternateRowColor}
\toprule
\multicolumn{3}{l}{{$\rightarrow \textit{Build on fully-connected classifier}$}} \\
\midrule

Xu~\cite{xu2023mimir} &   78.5   &    58.3    \\

Amini~\cite{amini2024meansparse} &   77.8   &    58.5    \\

Singh~\cite{singh2023revisiting} &   72.8   & 48.1  \\

Mo~\cite{mo2022adversarial} &   68.0  &   33.0    \\

\coloredMidrule{white}{white}
\toprule
\multicolumn{3}{l}{{$\rightarrow \textit{Build on distance-based classifier} $}} \\
\midrule

Xu~\cite{xu2023mimir} &  78.0  &  60.6(+\textbf{2.3}) \\

Amini~\cite{amini2024meansparse} & 76.1 & 60.5(+\textbf{2.0}) \\

Singh~\cite{singh2023revisiting} &   68.4  & 49.6(+\textbf{1.5}) \\

Mo~\cite{mo2022adversarial} &  67.7   & 35.7(+\textbf{2.7})  \\

\coloredBottomrule{alternateRowColor}
\end{tabular}

%% file: tabs/ablation_adv.tex
\rowcolors{3}{alternateRowColor}{white}
\begin{tabular}{ccc|cc}
\rowcolor{white}
\toprule
\bottomrule
\multicolumn{3}{c|}{\raisebox{-0.1em}{COMPONENTS}} & \multicolumn{2}{c}{\raisebox{-0.1em}{\textbf{Xu}~\cite{xu2023mimir}, ImageNet, $\ell_\infty=\nicefrac{4}{255}$}} \\

\raisebox{-0.3em}{CEest} & \raisebox{-0.3em}{PMixer} & \raisebox{-0.3em}{DistDeci} & \raisebox{-0.3ex}{~~~Clean Acc.~~~}  & \raisebox{-0.3ex}{Robust Acc.} \\
\midrule
\toprule
\multicolumn{3}{c|}{\textit{Default Classifier}}   & 78.5 & 58.3\\
\midrule
\checkmark &    \checkmark   &      \checkmark   & 78.0  &   60.6\textbf{(+2.3)}  \\

&   \checkmark   &     \checkmark     & 78.4 &   58.4\textbf{(+0.1)} \\
\checkmark&       &   \checkmark        & 78.2  &  58.8\textbf{(+0.5)}\\

&       &       \checkmark          &  76.5            & 55.1\textbf{(-3.2)} \\

\midrule
\checkmark &    \checkmark   &       \raisebox{-0.3ex}{\textit{FC-Head}}        &      73.1      & 55.3 \textbf{(-3.0)}\\

\coloredMidrule{white}{alternateRowColor}
\end{tabular}

%% file: tex/5_conclusion.tex
\section{Conclusion}\label{conclusion} 
In this paper, we analyzed the fundamental trade-off between sensitivity, discriminative power, and robustness in FC versus $\ell_2$-classifiers. To effectively leverage and unify their benefits, we proposed the Hybrid Prototype Mixing (HPM) framework, a dynamic $\ell_2$ re-classifier that fuses batch- and dataset-level prototypes. Recognizing that this architecture causes challenges like gradient obfuscation, we introduced the rigorous Mixed Surrogate Attack (MSA) protocol for fair evaluation. Our lightweight, plug-and-play HPM module successfully enhanced the adversarial robustness of various SOTA models with minimal fine-tuning.

%% file: supplementary.tex
\clearpage
\setcounter{page}{1}
\onecolumn
{
    \centering
    \Large
    \textbf{\thetitle}\\
    \vspace{0.5em}Supplementary Material \\
    \vspace{1.0em}

}

\section{Implementation Details}
\subsection{Environments}

Our project is implemented entirely in Python using PyTorch~\cite{NEURIPS2019_9015_pytorch} and PyTorch-Lightning~\cite{Falcon_PyTorch_Lightning_2019}, with all experiments conducted on a Linux server equipped with a single GPU. The versions and models of key software and hardware are summarized in Tab.~\ref{tab:env}.

\begin{table}[h] 
\centering
\resizebox{0.4\linewidth}{!}{
    \input{tabs/impl_env}
}
\caption{Key software versions and GPU model of our implementation. 
\textit{cu121} means the CUDA version is \textit{12.1}.}
\label{tab:env}
\end{table}

\subsection{Adversarial Fine-tuning}
When fine-tuning on clean data alone, we find it is often sufficient to enhance the robustness of certain models at some point. However, since these models primarily achieve adversarial robustness through adversarial training, we choose to further improve their performance using a similar adversarial training approach during fine-tuning. Specifically, we employ adversarial training based on the Improved Kullback-Leibler (IKL) loss~\cite{cui2305decoupled}. The IKL loss, an enhanced version of the Decoupled Kullback-Leibler (DKL) loss, is equivalent to the KL loss during gradient-based optimization and consists of two components: a weighted Mean Squared Error (wMSE) and a Cross-Entropy (CE) loss with soft labels. This composite loss serves as our optimization objective. During training, adversarial samples are generated to maximize the IKL loss over 10 gradient-based update steps\footnote{The implementation of the IKL loss and training procedure can be found in their official repository: \url{github.com/jiequancui/DKL}.}, and these samples are subsequently used for adversarial training. Additionally, we configure the key hyperparameters for the fine-tuning process as follows:

\begin{itemize}[leftmargin=*]
    \item The learning rate for Adam~\cite{kingma2014adam} optimizer is set to \(5 \times 10^{-5}\) for all models.
    \item  The momentum parameters \(\lambda_1\) and \(\lambda_2\) are set to \(\lambda_1 = \lambda_2 = 0.98\) for CIFAR-10, while for CIFAR-100 and ImageNet, they are increased to \(\lambda_1 = \lambda_2 = 0.995\) to accommodate more classes, given a fixed batch size of 64.
    
    \item For the CIFAR-10, the training amount is set to one epoch. This number is doubled due to the larger number of categories for CIFAR-100 and ImageNet.

\end{itemize}

\subsection{Evaluation Details} In this section , We detail our use of AutoAttack to evaluate the model's adversarial robustness under the two most powerful white-box attacks, specifically, APGD-CE, APGD-DLR~\cite{autoattack} and Square Attack~\cite{andriushchenko2020square}. Unlike the standard evaluation used in RobustBench~\cite{croce2020robustbench}, we omit the white-box FAB Attack~\cite{croce2020minimally}, as this attack is significantly less effective and fail to pose a threat to the models we evaluated. Therefore, similar to the setup in~\cite{bartoldson2024adversarial}\footnote{These setups can be found in their official implementation: \url{github.com/bbartoldson/Adversarial-Robustness-Limits}.}, we choose to exclude it.

Additionally, due to our limited computational resources, we further reduce the evaluation cost of adversarial robustness (as generating a single adversarial example involves dozens of forward and backward passes). First, we evaluate robustness using only 25\% of the test samples (for CIFAR-10 and CIFAR-100), which are uniformly sampled and then fixed. Second, we reduce the number of attack steps for generating adversarial examples from the default 100 steps to 40 (again, following~\cite{bartoldson2024adversarial}). This reduction is justified, as vulnerable samples are often successfully attacked within fewer steps, while increasing the number of steps for robust samples does not significantly improve the attack success rate. However, this setup inevitably allows some borderline samples to evade successful attacks. As a result, the overall accuracy we report (including that of baseline models) is approximately 0.5\% higher than the corresponding scores on the RobustBench~\cite{croce2020robustbench} leaderboard. Nevertheless, this discrepancy does not affect the direct comparisons between models in Tab.~\ref{table:adv_cifar} and Tab.~\ref{table:adv_imgnet} of the main paper, as all models follow exactly the same evaluation settings. Additionally, we list the concrete model architectures and their rankings on the RobustBench leaderboard for the evaluated works here:

\begin{itemize}
    \item Cui~\cite{cui2305decoupled} RANK\#~\textbf{No.9} on CIFAR-10, RANK\#~\textbf{No.4} on CIFAR-100; Architecture: \textbf{WideResNet-28-10}.
    \item Wang~\cite{wang2023better} RANK\#~\textbf{No.5} on CIFAR-10, RANK\#~\textbf{No.1} on CIFAR-100; Architecture: \textbf{WideResNet-70-16}.
    \item Jia~\cite{jia2022adversarial} RANK\#~\textbf{No.43} on CIFAR-10, RANK\#~\textbf{No.17} on CIFAR-100; Architecture: \textbf{WideResNet-70-16} for CIFAR-10 and \textbf{WideResNet-34-20} for CIFAR-100.
    \item Bartoldson~\cite{bartoldson2024adversarial} RANK\#~\textbf{No.1} on CIFAR-10, RANK\#~\textbf{N/A} on CIFAR-100; Architecture: \textbf{WideResNet-94-16}.
    \item Xu~\cite{xu2023mimir} RANK\#~\textbf{No.1} on ImageNet; Architecture: \textbf{Swin-L}.
    \item Amini~\cite{amini2024meansparse}  RANK\#~\textbf{No.6} on ImageNet; Architecture: \textbf{MeanSparse ConvNeXt-L}.
    \item Singh~\cite{singh2023revisiting} RANK\#~\textbf{No.16} on ImageNet; Architecture: \textbf{ConvNeXt-T + ConvStem}.
    \item Mo~\cite{mo2022adversarial} RANK\#~\textbf{No.26} on ImageNet; Architecture: \textbf{ViT-B}.
\end{itemize}

\section{Proof of Classifier Sensitivity}
\label{proof} We define the sensitivity of FC and DB classifiers in Eq.~\textcolor{red}{1} and Eq.~\textcolor{red}{2} of our main paper, reflecting how significantly the model’s predictions change in response to variations in the input feature representations. We first derive the supremum of $\mathbf{S}_d$ in Eq.~\textcolor{red}{5} of the main paper. The proof of this result is simply as follows:
\begin{align}
    \mathbf{S}_d &= |\frac{\|z_d + \epsilon_d -w_d\|- \|z_d-w_d\|}{\|z_d-w_d\|}|\leq \frac{\|\epsilon_d\|}{\|z_d-w_d\|} = \frac{\|\epsilon\|}{\sqrt{(\|z_d\|^2+\|w_d\|^2-2\|z_d\|\|w_d\|\cdot \mathrm{cos}(\theta_d))}}
\end{align}
Equivalence holds iff $\epsilon_d$ is collinear with $z_d - w_d$. Obviously, this value is the supremum of $\mathbf{S}_d$, which is the maximum influence on the model output when the feature changes. Then, we can easily derive the condition for $\mathbf{S}_c \geq \sup \mathbf{S}_d$ as follows:
\begin{align}
    \mathbf{S}_c \geq \mathrm{sup}~\mathbf{S}_d~ &\iff \frac{\|\epsilon_c\|\cdot |\mathrm{cos(\gamma)}|}{\|z_c\|\cdot |\mathrm{cos}(\theta_c)|} \geq \frac{\|\epsilon_d\|}{\sqrt{\|z_d\|^2+\|w_d\|^2-2\|z_d\|\|w_d\|\mathrm{cos}(\theta_d)}} \\
    &\iff ~\|z_c\|\cdot |\mathrm{cos}(\theta_c)| ~\leq |\mathrm{cos}(\gamma)|\cdot\sqrt{\|z_d\|^2+\|w_d\|^2-2\|z_d\|\|w_d\| \mathrm{cos}(\theta_d)}  \\
   &\iff~ |\mathrm{cos}(\theta_c)|\leq |\mathrm{cos}(\gamma)|\cdot \sqrt{\left(\frac{\|w_d\|}{\|z_c\|}\right)^2-2\cdot\mathrm{cos}(\theta_d)\frac{\|w_d\|}{\|z_c\|}+1} 
 \end{align}
 Meanwhile, we have $\sqrt{\left(\frac{\|w_d\|}{\|z\|}\right)^2-2\cdot\mathrm{cos}(\theta_d)\frac{\|w_d\|}{\|z\|}+1}\geq |\mathrm{sin}(\theta_d)|$, and then we can derive such condition as in Eq.~\textcolor{red}{8} of the main text, which indicates when FC classifiers are more sensitive to feature variations.
 
\subsection{Isotropic Distribution} Further more, we wish to derive the probability that such condition holds under our isotropic assumption. Firstly, we already know from~\cite{cai2013distributions} that the probability density functions of the random variables $\theta_d$ and $\theta_c$ are as follows:
 \begin{align}
     p(\theta,d) =\frac{\Gamma(\frac{d}{2})}{\Gamma(\frac{d-1}{2})\sqrt{\pi}}\cdot\mathrm{sin}^{d-2}\theta=\mathrm{C} \cdot \sin^{d-2}(\theta)
 \end{align}
where $\mathrm{C}$ is a coefficient term and $\theta \in [0,\pi]$. What we aim to explore is the distribution of $\nicefrac{|\mathrm{cos}(\theta_c)|}{|\mathrm{sin}(\theta_d)|}$. By performing a variable substitution: let $\eta=\cos(\theta)$ and $\tau=\sin(\theta)$, we can easily derive the distributions of \(\mathcal{H}=\cos(\theta_c)\) and \(\mathcal{T}= \sin(\theta_d)\), and their PDFs are as follows:
\begin{equation}
        p_\mathcal{H}(\eta, d)=\mathrm{C}\cdot(1-\eta^2)^{\frac{d-1}{2}-1},~ p_\mathcal{T}(\tau,d)= 2\mathrm{C}\cdot \frac{\tau^{d-2}}{\sqrt{1-\tau^2}} \label{eq:var_subst}
\end{equation}
Notably, when computing the distribution of \(\sin(\theta_d)\), attention should be paid to the range of \(\theta\). When making a variable substitution, we let \(\theta = \arcsin(\tau)\) where \(\tau \in [0,1]\), and then we should multiply the probability density by 2. Next, we can derive the distribution function of \(\mathcal{X}=|\nicefrac{\mathcal{H}}{\mathcal{T}}|\), and the calculation process is as follows:
\begin{align}
    F_\mathcal{X}(x)&=\mathrm{P}(\mathcal{X}\leq x)=\mathrm{P(|\mathcal{\frac{H}{T}}|\leq x)}=\mathrm{P}(|\mathcal{H}|\leq \mathcal{T}\cdot x) \\
    &=\int_{0}^1 p_\mathcal{T}(\tau, d)\cdot\mathrm{P}(|\mathcal{H}|\leq \tau\cdot x)~\mathrm{d}\tau \\
    &=\int_{0}^{1} p_\mathcal{T}(\tau,d)
    \int_{-\tau\cdot x}^{\tau\cdot x} p_\mathcal{H}(\eta,d)\mathrm{d}\eta\mathrm{d}\tau \label{eq:general_cdf}\\
    &=\int_0^1 2\mathrm{C}\cdot\frac{\tau^{d-2}}{\sqrt{1-\tau^2}}\int_0^{\tau \cdot x}2\mathrm{C}\cdot (1-\eta^2)^{\frac{d-3}{2}}\mathrm{d}\eta\mathrm{d}\tau
\end{align}
For the computational efficiency and simplicity, we let \(\tau = \sin(x)\), \(\eta = \sin(y)\) to further reparametrize integration variables in trigonometric form when numerically computing CDF as in Fig.~\textcolor{red}{2} of the main text, as follows:
\begin{equation}
    F_\mathcal{X}(z)= \frac{4\cdot\Gamma^2(\frac{d}{2})}{\pi\cdot\Gamma^2(\frac{d-1}{2})}\int_0^{\frac{\pi}{2}}\mathrm{sin}^{d-2}(x)\int_0^{\substack{\mathrm{sin}^{-1}(\\\mathrm{cos}(\gamma)\cdot\\\mathrm{sin}(z))}}\mathrm{cos}^{d-2}(y)\mathrm{d}x\mathrm{d}y
\end{equation}

\subsection{Empirical Distribution}
Here, we provide the proof hint of our conclusion in Eq.~\textcolor{red}{13} of the main text. First, based on our analysis and assumptions, the PDFs of $\theta_c$ and $\theta_d$ are given in Eq.~\textcolor{red}{12} of the main text. Then, we apply a variable substitution similar to Eq.~\eqref{eq:var_subst} to derive the PDFs of $p_\mathcal{H}$ and $p_\mathcal{T}$. Considering the complexity of their PDFs, however, we avoid applying variable substitution directly to their analytical forms. Instead, we derive a generalized form of $F_\mathcal{X}$ from Eq.\eqref{eq:general_cdf} first, which is easy to prove, for any $p_\mathcal{H}$ and $p_\mathcal{T}$ as follows:
\begin{align}
    F_{\mathcal{X}}(z)=\int_0^{1}\frac{p_{\theta_d}(\mathrm{sin^{-1}}(x))+p_{\theta_d}(\pi-\mathrm{sin^{-1}}(x))}{\sqrt{1-x^2}}\int_{-x\cdot z}^{\substack{x\cdot z}}\frac{p_{\theta_c}(\mathrm{cos^{-1}}(y))}{\sqrt{1-y^2}}\mathrm{d}x\mathrm{d}y
\end{align}
Finally, we apply simple variable substitution by letting $x=\sin(t)$ to simplify the results presented in Eq.\textcolor{red}{13} of the main text.


\section{Limitations and Future Work}
\label{limitation}
There are still several limitations in this work. First, we evaluate our model by fine-tuning pretrained models from existing works. However, since the fine-tuning setup may differ significantly from the original training configurations (e.g., training objectives, data augmentations), extensive fine-tuning could lead to overfitting and performance degradation, making it challenging to determine the optimal fine-tuning settings. Nonetheless, we see great potential in our method and believe that incorporating it into existing frameworks for from-scratch training could further enhance their robustness. Unfortunately, due to our very limited computational resources, we were unable to conduct such large-scale adversarial training. (For example, \cite{bartoldson2024adversarial} achieves SOTA performance on CIFAR-10 while requiring over $10^{21}$ training FLOPs!)
In addition, there remains considerable room for exploration regarding various design choices within our method, such as selecting better criterion functions (e.g., contrastive learning loss~\cite{mai2021scr_ncm}). As we aim to approach the problem from a simple perspective, specifically, by comparing FC classifiers and $\ell_2$-classifiers, and to provide a focused theoretical analysis, we are constrained by the scope and length of this paper and are therefore unable to explore the broader implications of classifier sensitivity differences from a higher-level perspective.
For instance, since this work reveals a link between feature sensitivity, adversarial robustness, and classification performance, a natural question arises: under a constraint that balances robustness and accuracy, can we derive an optimal form of classifier? Moreover, the sensitivity analysis presented in this paper remains at the level of individual logit elements. The quantitative effect of classifier sensitivity on the model’s overall prediction (particularly the top-confidence class) remains an open question. We hope these discussions can inspire future research.

%% file: tabs/impl_env.tex
\begin{tabular}{lcccl}
\toprule
\textbf{Component} & & & &\textbf{Version / Model} \\ 
\midrule
\midrule
System     &&&&  \textit{Ubuntu 22.04 LTS} \\ 
Python     &&&&  \textit{3.12.2} \\ 
PyTorch    &&&&  \textit{2.4.1+cu121} \\ 
GPU        &&&&  \textit{1$\times$ NVIDIA RTX 4090} \\ 
\bottomrule
\end{tabular}

%% file: main_bib.bib
@inproceedings{tsipras2019robustness,
  title={Robustness May Be at Odds with Accuracy},
  author={Tsipras, Dimitris and Santurkar, Shibani and Engstrom, Logan and Turner, Alexander and Madry, Aleksander},
  booktitle={International Conference on Learning Representations},
  number={2019},
  year={2019}
}

@inproceedings{athalye2018obfuscated,
  title={Obfuscated gradients give a false sense of security: Circumventing defenses to adversarial examples},
  author={Athalye, Anish and Carlini, Nicholas and Wagner, David},
  booktitle={International conference on machine learning},
  pages={274--283},
  year={2018},
  organization={PMLR}
}

@inproceedings{autoattack,
    title = {Reliable evaluation of adversarial robustness with an ensemble of diverse parameter-free attacks},
    author = {Francesco Croce and Matthias Hein},
    booktitle = {ICML},
    year = {2020}
}

@article{snell2017prototypical,
  title={Prototypical networks for few-shot learning},
  author={Snell, Jake and Swersky, Kevin and Zemel, Richard},
  journal={Advances in neural information processing systems},
  volume={30},
  year={2017}
}

@article{gogoi2022adaptive,
  title={Adaptive prototypical networks},
  author={Gogoi, Manas and Tiwari, Sambhavi and Verma, Shekhar},
  journal={arXiv preprint arXiv:2211.12479},
  year={2022}
}

@article{hodgson1988min_dis_classifier,
  title={Reducing the computational requirements of the minimum-distance classifier},
  author={Hodgson, Michael E},
  journal={Remote sensing of environment},
  volume={25},
  number={1},
  pages={117--128},
  year={1988},
  publisher={Elsevier}
}

@article{vinyals2016matching,
  title={Matching networks for one shot learning},
  author={Vinyals, Oriol and Blundell, Charles and Lillicrap, Timothy and Wierstra, Daan and others},
  journal={Advances in neural information processing systems},
  volume={29},
  year={2016}
}

@inproceedings{pan2019transferrable,
  title={Transferrable prototypical networks for unsupervised domain adaptation},
  author={Pan, Yingwei and Yao, Ting and Li, Yehao and Wang, Yu and Ngo, Chong-Wah and Mei, Tao},
  booktitle={Proceedings of the IEEE/CVF conference on computer vision and pattern recognition},
  pages={2239--2247},
  year={2019}
}

@article{ji2020improved,
  title={Improved prototypical networks for few-shot learning},
  author={Ji, Zhong and Chai, Xingliang and Yu, Yunlong and Pang, Yanwei and Zhang, Zhongfei},
  journal={Pattern Recognition Letters},
  volume={140},
  pages={81--87},
  year={2020},
  publisher={Elsevier}
}

@inproceedings{mai2021scr_ncm,
  title={Supervised contrastive replay: Revisiting the nearest class mean classifier in online class-incremental continual learning},
  author={Mai, Zheda and Li, Ruiwen and Kim, Hyunwoo and Sanner, Scott},
  booktitle={Proceedings of the IEEE/CVF conference on computer vision and pattern recognition},
  pages={3589--3599},
  year={2021}
}

@inproceedings{beyer1999nearest,
  title={When is “nearest neighbor” meaningful?},
  author={Beyer, Kevin and Goldstein, Jonathan and Ramakrishnan, Raghu and Shaft, Uri},
  booktitle={Database Theory—ICDT’99: 7th International Conference Jerusalem, Israel, January 10--12, 1999 Proceedings 7},
  pages={217--235},
  year={1999},
  organization={Springer}
}

@inproceedings{gao2019hybrid,
  title={Hybrid attention-based prototypical networks for noisy few-shot relation classification},
  author={Gao, Tianyu and Han, Xu and Liu, Zhiyuan and Sun, Maosong},
  booktitle={Proceedings of the AAAI conference on artificial intelligence},
  volume={33},
  number={01},
  pages={6407--6414},
  year={2019}
}

@article{mettes2019hyperspherical,
  title={Hyperspherical prototype networks},
  author={Mettes, Pascal and Van der Pol, Elise and Snoek, Cees},
  journal={Advances in neural information processing systems},
  volume={32},
  year={2019}
}

@article{ben2006analysis,
  title={Analysis of representations for domain adaptation},
  author={Ben-David, Shai and Blitzer, John and Crammer, Koby and Pereira, Fernando},
  journal={Advances in neural information processing systems},
  volume={19},
  year={2006}
}

@article{parisi2019continual,
  title={Continual lifelong learning with neural networks: A review},
  author={Parisi, German I and Kemker, Ronald and Part, Jose L and Kanan, Christopher and Wermter, Stefan},
  journal={Neural networks},
  volume={113},
  pages={54--71},
  year={2019},
  publisher={Elsevier}
}

@article{guerriero2018deepncm,
  title={Deepncm: Deep nearest class mean classifiers},
  author={Guerriero, Samantha and Caputo, Barbara and Mensink, Thomas},
  year={2018}
}

@article{cover1967nearest,
  title={Nearest neighbor pattern classification},
  author={Cover, Thomas and Hart, Peter},
  journal={IEEE transactions on information theory},
  volume={13},
  number={1},
  pages={21--27},
  year={1967},
  publisher={IEEE}
}

@article{wang2022visual,
  title={Visual recognition with deep nearest centroids},
  author={Wang, Wenguan and Han, Cheng and Zhou, Tianfei and Liu, Dongfang},
  journal={arXiv preprint arXiv:2209.07383},
  year={2022}
}

@inproceedings{deng2009imagenet,
  title={Imagenet: A large-scale hierarchical image database},
  author={Deng, Jia and Dong, Wei and Socher, Richard and Li, Li-Jia and Li, Kai and Fei-Fei, Li},
  booktitle={2009 IEEE conference on computer vision and pattern recognition},
  pages={248--255},
  year={2009},
  organization={Ieee}
}

@article{krizhevsky2009cifar,
  title={Cifar-10 and cifar-100 datasets},
  author={Krizhevsky, Alex and Nair, Vinod and Hinton, Geoffrey},
  journal={URl: https://www.cs.toronto.edu/~kriz/cifar.html},
  volume={6},
  number={1},
  pages={1},
  year={2009}
}

@article{bartoldson2024adversarial,
  title={Adversarial Robustness Limits via Scaling-Law and Human-Alignment Studies},
  author={Bartoldson, Brian R and Diffenderfer, James and Parasyris, Konstantinos and Kailkhura, Bhavya},
  journal={arXiv preprint arXiv:2404.09349},
  year={2024}
}

@article{croce2020robustbench,
  title={Robustbench: a standardized adversarial robustness benchmark},
  author={Croce, Francesco and Andriushchenko, Maksym and Sehwag, Vikash and Debenedetti, Edoardo and Flammarion, Nicolas and Chiang, Mung and Mittal, Prateek and Hein, Matthias},
  journal={arXiv preprint arXiv:2010.09670},
  year={2020}
}

@inproceedings{he2016deep,
  title={Deep residual learning for image recognition},
  author={He, Kaiming and Zhang, Xiangyu and Ren, Shaoqing and Sun, Jian},
  booktitle={Proceedings of the IEEE conference on computer vision and pattern recognition},
  pages={770--778},
  year={2016}
}

@article{wang2020generalizing,
  title={Generalizing from a few examples: A survey on few-shot learning},
  author={Wang, Yaqing and Yao, Quanming and Kwok, James T and Ni, Lionel M},
  journal={ACM computing surveys (csur)},
  volume={53},
  number={3},
  pages={1--34},
  year={2020},
  publisher={ACM New York, NY, USA}
}

@inproceedings{romera2015embarrassingly,
  title={An embarrassingly simple approach to zero-shot learning},
  author={Romera-Paredes, Bernardino and Torr, Philip},
  booktitle={International conference on machine learning},
  pages={2152--2161},
  year={2015},
  organization={PMLR}
}

@article{michel2022survey_vulnerale,
  title={A survey on the vulnerability of deep neural networks against adversarial attacks},
  author={Michel, Andy and Jha, Sumit Kumar and Ewetz, Rickard},
  journal={Progress in Artificial Intelligence},
  volume={11},
  number={2},
  pages={131--141},
  year={2022},
  publisher={Springer}
}

@article{bortsova2021adversarial_vulnerable,
  title={Adversarial attack vulnerability of medical image analysis systems: Unexplored factors},
  author={Bortsova, Gerda and Gonz{\'a}lez-Gonzalo, Cristina and Wetstein, Suzanne C and Dubost, Florian and Katramados, Ioannis and Hogeweg, Laurens and Liefers, Bart and van Ginneken, Bram and Pluim, Josien PW and Veta, Mitko and others},
  journal={Medical Image Analysis},
  volume={73},
  pages={102141},
  year={2021},
  publisher={Elsevier}
}

@article{szegedy2013intriguing,
  title={Intriguing properties of neural networks},
  author={Szegedy, C},
  journal={arXiv preprint arXiv:1312.6199},
  year={2013}
}

@article{goodfellow2014explaining,
  title={Explaining and harnessing adversarial examples},
  author={Goodfellow, Ian J},
  journal={arXiv preprint arXiv:1412.6572},
  year={2014}
}

@article{madry2017towards,
  title={Towards deep learning models resistant to adversarial attacks},
  author={Madry, Aleksander},
  journal={arXiv preprint arXiv:1706.06083},
  year={2017}
}

@article{mensink2013distance,
  title={Distance-based image classification: Generalizing to new classes at near-zero cost},
  author={Mensink, Thomas and Verbeek, Jakob and Perronnin, Florent and Csurka, Gabriela},
  journal={IEEE transactions on pattern analysis and machine intelligence},
  volume={35},
  number={11},
  pages={2624--2637},
  year={2013},
  publisher={IEEE}
}

@article{ba2016layer,
  title={Layer normalization},
  author={Ba, Jimmy Lei},
  journal={arXiv preprint arXiv:1607.06450},
  year={2016}
}

@article{cui2305decoupled,
  title={Decoupled kullback-leibler divergence loss},
  author={Cui, Jiequan and Tian, Zhuotao and Zhong, Zhisheng and Qi, Xiaojuan and Yu, Bei and Zhang, Hanwang},
  journal={Advances in Neural Information Processing Systems},
  volume={37},
  pages={74461--74486},
  year={2024}
}

@inproceedings{wang2023better,
  title={Better diffusion models further improve adversarial training},
  author={Wang, Zekai and Pang, Tianyu and Du, Chao and Lin, Min and Liu, Weiwei and Yan, Shuicheng},
  booktitle={International Conference on Machine Learning},
  pages={36246--36263},
  year={2023},
  organization={PMLR}
}

@inproceedings{jia2022adversarial,
  title={LAS-AT: adversarial training with learnable attack strategy},
  author={Jia, Xiaojun and Zhang, Yong and Wu, Baoyuan and Ma, Ke and Wang, Jue and Cao, Xiaochun},
  booktitle={Proceedings of the IEEE/CVF Conference on Computer Vision and Pattern Recognition},
  pages={13398--13408},
  year={2022}
}

@book{webb2003statistical,
  title={Statistical pattern recognition},
  author={Webb, Andrew R},
  year={2003},
  publisher={John Wiley \& Sons}
}

@incollection{NEURIPS2019_9015_pytorch,
title = {PyTorch: An Imperative Style, High-Performance Deep Learning Library},
author = {Paszke, Adam and Gross, Sam and Massa, Francisco and Lerer, Adam and Bradbury, James and Chanan, Gregory and Killeen, Trevor and Lin, Zeming and Gimelshein, Natalia and Antiga, Luca and Desmaison, Alban and Kopf, Andreas and Yang, Edward and DeVito, Zachary and Raison, Martin and Tejani, Alykhan and Chilamkurthy, Sasank and Steiner, Benoit and Fang, Lu and Bai, Junjie and Chintala, Soumith},
booktitle = {Advances in Neural Information Processing Systems 32},
editor = {H. Wallach and H. Larochelle and A. Beygelzimer and F. d\textquotesingle Alch\'{e}-Buc and E. Fox and R. Garnett},
pages = {8024--8035},
year = {2019},
publisher = {Curran Associates, Inc.},
url = {http://papers.neurips.cc/paper/9015-pytorch-an-imperative-style-high-performance-deep-learning-library.pdf}
}

@inproceedings{croce2020minimally,
  title={Minimally distorted adversarial examples with a fast adaptive boundary attack},
  author={Croce, Francesco and Hein, Matthias},
  booktitle={International Conference on Machine Learning},
  pages={2196--2205},
  year={2020},
  organization={PMLR}
}

@inproceedings{andriushchenko2020square,
  title={Square attack: a query-efficient black-box adversarial attack via random search},
  author={Andriushchenko, Maksym and Croce, Francesco and Flammarion, Nicolas and Hein, Matthias},
  booktitle={European conference on computer vision},
  pages={484--501},
  year={2020},
  organization={Springer}
}

@article{kingma2014adam,
  title={Adam: A method for stochastic optimization},
  author={Kingma, Diederik P},
  journal={arXiv preprint arXiv:1412.6980},
  year={2014}
}

@article{cai2013distributions,
  title={Distributions of angles in random packing on spheres},
  author={Cai, Tony and Fan, Jianqing and Jiang, Tiefeng},
  journal={The Journal of Machine Learning Research},
  volume={14},
  number={1},
  pages={1837--1864},
  year={2013},
  publisher={JMLR. org}
}

@article{xu2023mimir,
  title={MIMIR: Masked Image Modeling for Mutual Information-based Adversarial Robustness},
  author={Xu, Xiaoyun and Yu, Shujian and Liu, Zhuoran and Picek, Stjepan},
  journal={arXiv preprint arXiv:2312.04960},
  year={2023}
}

@article{singh2023revisiting,
  title={Revisiting adversarial training for imagenet: Architectures, training and generalization across threat models},
  author={Singh, Naman Deep and Croce, Francesco and Hein, Matthias},
  journal={Advances in Neural Information Processing Systems},
  volume={36},
  pages={13931--13955},
  year={2023}
}

@article{mo2022adversarial,
  title={When adversarial training meets vision transformers: Recipes from training to architecture},
  author={Mo, Yichuan and Wu, Dongxian and Wang, Yifei and Guo, Yiwen and Wang, Yisen},
  journal={Advances in Neural Information Processing Systems},
  volume={35},
  pages={18599--18611},
  year={2022}
}

@article{amini2024meansparse,
  title={MeanSparse: Post-training robustness enhancement through mean-centered feature sparsification},
  author={Amini, Sajjad and Teymoorianfard, Mohammadreza and Ma, Shiqing and Houmansadr, Amir},
  journal={arXiv preprint arXiv:2406.05927},
  year={2024}
}

@article{ilyas2019adversarial,
  title={Adversarial examples are not bugs, they are features},
  author={Ilyas, Andrew and Santurkar, Shibani and Tsipras, Dimitris and Engstrom, Logan and Tran, Brandon and Madry, Aleksander},
  journal={Advances in neural information processing systems},
  volume={32},
  year={2019}
}

@article{mao2019metric,
  title={Metric learning for adversarial robustness},
  author={Mao, Chengzhi and Zhong, Ziyuan and Yang, Junfeng and Vondrick, Carl and Ray, Baishakhi},
  journal={Advances in neural information processing systems},
  volume={32},
  year={2019}
}

@inproceedings{musgrave2020metric,
  title={A metric learning reality check},
  author={Musgrave, Kevin and Belongie, Serge and Lim, Ser-Nam},
  booktitle={European Conference on Computer Vision},
  pages={681--699},
  year={2020},
  organization={Springer}
}

@inproceedings{schroff2015facenet,
  title={Facenet: A unified embedding for face recognition and clustering},
  author={Schroff, Florian and Kalenichenko, Dmitry and Philbin, James},
  booktitle={Proceedings of the IEEE conference on computer vision and pattern recognition},
  pages={815--823},
  year={2015}
}

@article{kaya2019deep,
  title={Deep metric learning: A survey},
  author={Kaya, Mahmut and Bilge, Hasan {\c{S}}akir},
  journal={Symmetry},
  volume={11},
  number={9},
  pages={1066},
  year={2019},
  publisher={MDPI}
}

@inproceedings{hoffer2015deep,
  title={Deep metric learning using triplet network},
  author={Hoffer, Elad and Ailon, Nir},
  booktitle={International workshop on similarity-based pattern recognition},
  pages={84--92},
  year={2015},
  organization={Springer}
}

@inproceedings{papernot2017practical,
  title={Practical black-box attacks against machine learning},
  author={Papernot, Nicolas and McDaniel, Patrick and Goodfellow, Ian and Jha, Somesh and Celik, Z Berkay and Swami, Ananthram},
  booktitle={Proceedings of the 2017 ACM on Asia conference on computer and communications security},
  pages={506--519},
  year={2017}
}

@inproceedings{chen2017zoo,
  title={Zoo: Zeroth order optimization based black-box attacks to deep neural networks without training substitute models},
  author={Chen, Pin-Yu and Zhang, Huan and Sharma, Yash and Yi, Jinfeng and Hsieh, Cho-Jui},
  booktitle={Proceedings of the 10th ACM workshop on artificial intelligence and security},
  pages={15--26},
  year={2017}
}

@article{weinberger2009distance,
  title={Distance metric learning for large margin nearest neighbor classification.},
  author={Weinberger, Kilian Q and Saul, Lawrence K},
  journal={Journal of machine learning research},
  volume={10},
  number={2},
  year={2009}
}

@inproceedings{aggarwal2001surprising,
  title={On the surprising behavior of distance metrics in high dimensional space},
  author={Aggarwal, Charu C and Hinneburg, Alexander and Keim, Daniel A},
  booktitle={International conference on database theory},
  pages={420--434},
  year={2001},
  organization={Springer}
}

@misc{papaspiliopoulos2020high,
  title={High-dimensional probability: An introduction with applications in data science},
  author={Papaspiliopoulos, Omiros},
  year={2020},
  publisher={Taylor \& Francis}
}

@article{bai2021recent,
  title={Recent advances in adversarial training for adversarial robustness},
  author={Bai, Tao and Luo, Jinqi and Zhao, Jun and Wen, Bihan and Wang, Qian},
  journal={arXiv preprint arXiv:2102.01356},
  year={2021}
}

@software{Falcon_PyTorch_Lightning_2019,
author = {Falcon, William and {The PyTorch Lightning team}},
doi = {10.5281/zenodo.3828935},
license = {Apache-2.0},
month = mar,
title = {{PyTorch Lightning}},
url = {https://github.com/Lightning-AI/lightning},
version = {1.4},
year = {2019}
}
